\documentclass{article}
\usepackage{arxiv}
\usepackage[utf8]{inputenc} 
\usepackage[T1]{fontenc}    
\usepackage{hyperref}       
\usepackage{url}            
\usepackage{booktabs}       
\usepackage{amsfonts}       
\usepackage{nicefrac}       
\usepackage{microtype}      
\usepackage{lipsum}		
\usepackage{graphicx}
\usepackage{amsmath,amssymb,amsfonts}
\usepackage{algorithmic}
\usepackage{graphicx}
\usepackage{textcomp}
\usepackage{xcolor}
\usepackage{booktabs} 
\usepackage{graphicx}
\usepackage{epstopdf}

\usepackage{url}
\usepackage{algorithmic}
\usepackage{hyperref}
\usepackage{multirow}
\usepackage{caption}
\usepackage{amssymb}
\usepackage{amsfonts}
\usepackage{float}
\usepackage{bm}
\usepackage{dsfont}
\usepackage[T1]{fontenc}
\usepackage[left,modulo]{lineno}
\usepackage[ruled, vlined, linesnumbered, longend]{algorithm2e}
\usepackage{psfrag}
\usepackage{subfigure}
\usepackage{fancyhdr}
\usepackage{eucal}
\usepackage[english]{babel}
\usepackage{cite}

\usepackage{ifpdf}
\usepackage{setspace}
\usepackage{textcomp}
\usepackage{tikz}
\usepackage{tikz-inet}
\usetikzlibrary{calc,shapes.geometric}
\usetikzlibrary{shapes,snakes,arrows,automata} 
\usetikzlibrary{patterns}

\newif\ifnotes\notestrue

\newcommand{\ben}{\begin{enumerate}}
\newcommand{\een}{\end{enumerate}}

\newcommand{\bc}{\begin{center}}
\newcommand{\ec}{\end{center}}

\newcommand{\bit}{\begin{itemize}}
\newcommand{\eit}{\end{itemize}}

\newcommand{\ds}{\displaystyle}
\newcommand{\beq}{\begin{equation}}
\newcommand{\eeq}{\end{equation}}

\newcommand{\wre}{\mathbf{W}^{\rm{h}}}
\newcommand{\wi}{\mathbf{W}^{\rm{in}}}
\newcommand{\wo}{\mathbf{W}^{\rm{out}}}
\newcommand{\wfb}{\mathbf{W}^{\rm{fb}}}

\newcommand{\va}{\mathbf{u}}

\newcommand{\x}{\mathbf{x}}
\newcommand{\y}{\mathbf{y}}

\newcommand{\Na}{n}

\newcommand{\R}{\mathds{R}}

\renewcommand{\wre}{\mathbf{W}^{\rm{r}}}
\renewcommand{\wfb}{\mathbf{W}^{\rm{fb}}}

\newcommand{\noise}{\bm{\varepsilon}}
\newcommand{\idct}{\phi}

\renewcommand{\Na}{K}
\newcommand{\Nx}{N}
\newcommand{\Ny}{L}

\newif\ifnotes\notestrue

\def\hgr#1{}

\newcommand{\ti}{\!\times\!}
\newcommand{\dimwa}{\Nx\ti\Na}
\newcommand{\dimwr}{\Nx\ti\Nx}
\newcommand{\dimwfb}{\Nx\ti\Ny}
\newcommand{\dimwo}{\Nx\ti(\Na\!+\!\Ny)}

\newcommand{\espacio}[1]{}

\title{Evolutionary Echo State Network:\\
evolving reservoirs in the Fourier space}

\author{\href{https://orcid.org/0000-0002-9172-0155}{\includegraphics[scale=0.06]{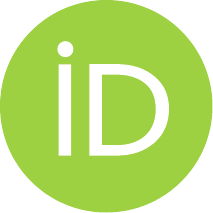}\hspace{1mm}Sebasti\'an Basterrech}\\
	Faculty of Electrical Engineering and Computer Science\\
	V\v{S}B-Technical University of Ostrava, Ostrava, Czech Republic \\
	\texttt{Sebastian.Basterrech@vsb.cz} \\
	\And
	\href{https://orcid.org/0000-0002-1712-0477}{\includegraphics[scale=0.06]{orcid.pdf}\hspace{1mm}Gerardo Rubino} \\
	INRIA Rennes -- Bretagne Atlantique\\
	Rennes, France\\
	\texttt{Gerardo.Rubino@inria.fr}\\
}

\date{}
\renewcommand{\headeright}{To appear in IJCNN'2022}
\renewcommand{\undertitle}{(Accepted manuscript at IJCNN'2022)}
\renewcommand{\shorttitle}{Evolutionary Echo State Networks}

\hypersetup{
pdftitle={Evolutionary Echo State Network: evolving reservoirs in the Fourier space},
pdfsubject={cs.NE,cs.AI,cs.LG},
pdfauthor={Sebasti\'an Basterrech, Gerardo Rubino},
pdfkeywords={Recurrent Neural Networks, Echo State Networks, Genetic Algorithms, Reservoir Computing, Fourier Transform},
}

\begin{document}
\maketitle

\begin{abstract}
The Echo State Network (ESN) is a class of Recurrent Neural Network with a large number of hidden-hidden weights (in the so-called reservoir).
%
Canonical ESN and its variations have recently received significant attention due to their remarkable success in the modeling of non-linear dynamical systems.
The reservoir is randomly connected with fixed weights that don't change in the learning process. Only the weights from reservoir to output are trained. Since the reservoir is fixed during the training procedure, 
we may wonder if the computational power of the recurrent structure is fully harnessed.

%
%

%
In this article, we propose a new computational model of the ESN type, that represents the reservoir weights in the Fourier space and performs a fine-tuning of these weights applying genetic algorithms in the frequency domain.
The main interest is that this procedure will work in a much smaller space compared to the classical ESN, thus providing a dimensionality reduction transformation of the initial method.
The proposed technique allows us to exploit the benefits of the large recurrent structure avoiding the training problems of gradient-based method.
We provide a detailed experimental study that demonstrates the good performances of
our approach with well-known chaotic systems and real-world data.
\end{abstract}

\keywords{Recurrent Neural Networks \and  Echo State Networks \and Genetic Algorithms \and Reservoir Computing \and Fourier Transform}

\section{Introduction}
\label{Introduction}
A Neural Network (NN) with a recurrent topology (RNN) is a powerful tool for modeling time series and solving machine learning problems with sequential data. 
Recurrences (circuits) are relevant for the computational power of neural architectures; thanks to them, hidden states are able to store more information from input streams.
Numerous works explain the distinct advantages presented by RNNs~\cite{Martens2011,Tanaka2019,Schmidhuber15,Jaeger09}: good theoretical properties, biologically plausibility, success in several real-world applications.
%
%
%
%
%
%
%
%
However, it is still hard to properly define optimal network architectures in RNNs~\cite{Tanaka2019}.
The difficulties are even higher when the network is large and the input signal contains long-term data dependencies~\cite{Martens2011}.
In summary, the principal well-identified problems are: not guaranteed convergence, long training times, and difficulties in storing information in internal states for long time periods (long-range memory)~\cite{Jaeger09,Martens2011,Schmidhuber15}.

A class of single-layer RNNs called Reservoir Computing (RC) has attracted considerable attention during the last years due to its remarkable success in the modeling of complex non-linear dynamical systems~\cite{Tanaka2019}.
%
%
%
An RC model relies on the assumption that supervised adaptation of all hidden-hidden weights is not necessary~\cite{Jaeger09}.
A single-layer recurrent network (so-called \textit{reservoir}) is driven by an input signal, and projects the input information into a high-dimensional space.
Reservoir weights are randomly initialized and scaled for satisfying technical conditions of dynamic stability~\cite{Jaeger09}.
A distinctive characteristic of an RC model is that the training algorithm only adjusts the readout structure, in general recurrence-free weights that map the reservoir projection to the output space. Nowadays, learning algorithms for adapting the recurrence-free parameters are fast and robust.
%
%

%
The research community has been actively studying methods for creating efficient reservoir models~\cite{Tanaka2019}.
Specific cyclic topologies were studied in~\cite{Rodan2012,Rodan11}, and pre-training techniques with self-organized strategies were analyzed in~\cite{MantasThesis,BasterCord11}. 
%
%
In~\cite{Schrauwen07, Steil2007}, reservoir units with parametric activation functions were trained for maximizing the entropy using the Intrinsic Plasticity (IP) rule.
Other approaches based on Evolutionary Algorithms (EAs) for optimizing hyper-parameters of the reservoir structure, such as number of reservoir units, scaling parameters, connectivity properties (density and spectral radius of the reservoir weights matrix), have also been introduced in the literature~\cite{Ferreira11,Ferreira2013}.
EAs have also been used for optimizing interconnected reservoirs~\cite{Dale2018,Qianli2020}, where GAs are applied for searching hyper-parameters of hierarchical reservoirs.
Besides, Bayesian optimization for finding  hyper-parameters of particular reservoirs was also analyzed in~\cite{Maat2018,Ribeiro2020}.
Furthermore, a relatively new prominent research in the RC area consists of models with interconnected reservoirs.
A tree structure where the tree-nodes satisfies the reservoir characteristics was developed in~\cite{GallicchioTree}. 
The weight connections among the reservoirs are set in order of creating a global contractive mapping.
Furthermore, layered/hierarchical architectures composed of multi-layered reservoirs have also been studied in~\cite{Qianli2020,Gallicchio2019b,Gallicchio2019, Dale2018,Gallicchio2017,GALLICCHIO201787}.

\medskip
\noindent\textbf{Motivations.}
Defining good hyper-parameters of ESNs is still often made by experience using a grid search strategy, partially brute force, and in general requires numerous trails~\cite{Rodan2012}.
%
%
Another drawback is that different reservoir matrices with the same hyper-parameters may produce substantially different results~\cite{Schrauwen07}. 
%
%
In addition, since the reservoir is fixed during the training procedure, the computational power of the model is not fully harnessed~\cite{Martens2011,Rodan2012}.

\medskip
\noindent\textbf{Contributions.}
In this article, we introduce the EVOlutionary Echo State Network (EvoESN), a novel computational model rooted in RC concepts.
We combine ideas regarding dimensionality reduction using the Fourier transform for searching weights of RNNs (fully connected architectures)~\cite{Koutnik2010}, optimization of network weights using an evolutionary search (such as in EVOLINO~\cite{Schmidhuber2007}), and the potential of ESNs for fast and robust temporal learning~\cite{Jaeger09}.
%
%

Given a reservoir with a fixed pattern of connectivity (as in standard ESNs), we apply an indirect encoding procedure for representing the reservoir weights into the Fourier space (specifically, by means of the Discrete Cosine Transform, DCT)~\cite{Koutnik2010,Koutnik2013}. 
Then, we perform a fine-tuning of the reservoir weights applying genetic algorithms over the DCT coefficients.
It is important to note that we are not looking for a good reservoir directly working with the weights, that becomes changing the RC paradigm and leaving the fast learning advantages of an RC model.
Instead, we solve an optimization problem in the frequency domain. Using this Fourier representation drastically reduces the size of the searching space, and makes it possible to efficiently look for good reservoir weights.
Our framework is a strong integration between learning in the RC paradigm and using EAs to improve the reservoir (to find \textit{better} weights) following a global iterative process moving at each iteration between the weights space and the frequency space using the DCT transform and its inverse.
%

%
%
%
We demonstrate the utility of EvoESN by experiments on two well-known chaotic systems (the Mackey-Glass system and the Lorenz systems~\cite{JaegerScience04,Schmidhuber2007}) and a real-world problem (Sunspot series~\cite{sidc}). 
Preliminary results of part of the content of this paper were briefly presented in the 2-page poster~\cite{BasterGECCO21}.

The rest of the article is organized as follows. In the next section we present the ESN framework. Our main contributions are the object of Section~\ref{Methodology}, where we discuss the reservoir encoding  and the evolutionary search approach.
Experimental results are provided in Section~\ref{ExperimentalResults}.
The article ends in Section~\ref{Conclusions} with a discussion about the presented work, its limitations and possible future research avenues.

\section{Preliminaries}
\subsection{The Echo State Network framework}
\label{RC}

The large recognition of ESNs comes from their record-breaking success in predicting observations through the Mackey-Glass attractor~\cite{JaegerScience04}.
The most common application of ESNs is time series prediction, where the operation is divided into two phases. An initial training phase where the model is externally driven by an input signal and the parameters are trained used supervised learning, followed by an autonomous phase (so-called free-run prediction) where the input signal is provided by the model itself using previous predicted values~\cite{Hart2020}.

Let us consider an ESN with $\Na$ input neurons, $\Nx > \Na$ hidden neurons and $\Ny$ output neurons.
The dynamics of the model is described in terms of a discrete-time system with an input signal $\va(t)\in \R^{\Na}$ and a hidden state $\x(t)\in \R^{\Nx}$. The recurrent state is computed according to the following recurrence~\cite{Jaeger01}:
\begin{equation}
\label{hiddenStateFeedback}
\x(t)=f\big(\wi\va(t)+\wre\x(t-1)+\wfb \y(t-1)+\noise(t)\big),
\end{equation}
where~$t$ is the discrete time index, $\wi$ is a $\dimwa$ matrix with input-to-reservoir weights, the $\dimwr$ matrix $\wre$ has the reservoir internal weights, $\wfb$ is a weight $\dimwfb$ matrix with the feedback connections going from the output neurons to the reservoir, and $\noise(t)$ is a noise vector with dimension $\Nx$.
Function $f(\cdot)$ is the activation function, usually an hyperbolic function, applied element-wise.
Matrix $\wfb$ and vector $\noise$ are optional~\cite{JaegerScience04}.
%
%
The model's output $\y(t)\in\R^{\Ny}$ is computed by a readout layer defined as:
\begin{equation}
\label{outputState}
\y(t)=g(\wo[\va(t);\x(t)]),
\end{equation}
where $\wo$ is the $\dimwo$ reservoir-to-output weight matrix, $g(\cdot)$ is a coordinate-wise function (usually an hyperbolic tangent or, in some cases, the identity) and $[\cdot;\cdot]$ denotes vector concatenation.
The readout weights are trained using a classical tool such as Tikhonov regularization~\cite{Jaeger09,Hart2020}.
In the case of an ESN with feedback connections (non-null $\wfb$), during the training procedure the feedback values are \textit{teacher-forced}: instead of feeding back the reservoir with the predicted value $\y(t)$, the real target is used as the feedback value~\cite{Felix2012}. In exploitation mode, the model's output $\y(t)$ is fed back to the reservoir.

Since the reservoir weights are frozen during learning, the initial reservoir design may have an impact on the model's performance.
The most relevant hyper-parameters to be chosen are: reservoir size, input scaling factor, activation functions and reservoir spectral radius.  
Most research effort has gone into understanding the impact of those parameters. 
For more information, especially about the fundamental property of the reservoir dynamics (so-called \textit{Echo State Property} (ESP)), we suggest~\cite{Hart2020,MantasPracticalGuide12,Jaeger09,Tanaka2019,BasterrechIJCNN2017}.

\subsection{Evolutionary Algorithms applied on RC models}
There has been a rising interest in large-scale neural architecture engineering using evolutionary algorithms~\cite{Elsken2019}. 
Several EAs have been developed to optimize NNs' architectures~\cite{Schmidhuber2007,Rawal2016,StanleyNature2019}.
The RC field has also taken advantage of the recent advances in evolutionary computation.
%
%
EAs have been employed for finding the hyper-parameters of an RC model, such as: sparsity, input scaling factor, injected noise, spectral radius, leaky rate and number of reservoir units.
Particle Swarm Optimization (PSO) and Genetic Algorithms (GAs) have been used as metaheuristics for selecting optimal hyperparameters in~\cite{Anderson12,Ferreira11,Ferreira2013}. 
Global parameters of an extended ESN (DRESN) were optimized using GAs for solving medical signal prognosis~\cite{Zhong2017}.
EAs have also been applied for evolving the topology of some RC models.
An evolutionary pre-training procedure based on the extreme learning approach has been studied in~\cite{BasterrechNabic14}, where the authors applied PSO for adjusting a subset of the reservoir weights.
Recently, NEAT and HyperNEAT algorithms were also applied for evolving the pattern of connectivity of reservoirs in~\cite{Chatzidimitriou2010,Matzner2017}.
In~\cite{Matzner2017} the evolution of the reservoir topology was made considering a fitness function that combines the memory capacities of the model and its accuracy.
HyperNEAT algorithm over a full recurrent topology has a significant computational cost~\cite{Matzner2017}.
GAs have also been used for optimizing the hyper-parameters of hierarchical reservoirs and for finding connection parameters between reservoir structures~\cite{Qianli2020,Dale2018}.
Furthermore, a microbial GA was used for optimizing multilayered architectures of interconnected reservoirs~\cite{Dale2018}. 
It was also employed for finding good architectures of single ESNs with leaky integrator units.
Recently, a hybrid approach using swarm optimization and local search was applied for evolving hierarchical ESNs~\cite{Long2020}.

\medskip
\noindent\textbf{Differences between EvoESN and previous works.} 
There are also other works that tune the reservoir weights using evolutionary techniques.
%
%
We identify two relevant differences between our approach and the proposed in~\cite{Matzner2017}.
EvoESN performs evolution over a fixed reservoir topology and the fine-tuning is done only over the weights, while in~\cite{Matzner2017} the evolution is applied in both structure and connection weights of the reservoirs using HyperNEAT.
In addition, we perform the optimization of the weights in the Fourier space applying a considerable reduction of the searching space.
%
As already mentioned above, there are also other works where was applied EA over the reservoir weights, e.g. microbial GA was used on a single ESN~\cite{Dale2018} and PSO in~\cite{BasterrechNabic14}. 
The difference between EvoESN  and these previous works is the encoding technique, in the mentioned works the evolution was applied directly over the weight connections.


\section{EvoESN: Evolutionary Echo State Network}\label{Methodology}
%
\subsection{Dimensionality reduction
of reservoir weights applying a Fourier-type transform}
\label{Encoding}
%
%

%

\newcommand{\Nc}{C}
\newcommand{\coeff}{\bm{\alpha}}
%
%
%
%
%
%
%

Genetic encoding is a fundamental aspect of EAs, with special interest when  algorithms are applied for optimizing large complex structures.
A direct encoding (that is, a one-to-one mapping where every weight is represented in the gene) is clearly unsuitable for large-scale networks~\cite{Koutnik2013}.
Furthermore, it has been shown that naive crossover operations present deficiencies when used for combining recurrent networks~\cite{Lehman2018}.
%
%
%
In~\cite{Koutnik2010}, an indirect encoding procedure for NNs was introduced, where the network weights are represented in the frequency domain applying a Fourier transform, more precisely, the type-II DCT of the weight matrix.
The evolutionary search is conducted in a reduced space composed by the coefficients of the low frequencies only~\cite{Koutnik2013}.
This leads to a significant dimensionality reduction.
%
%
%
%
The approach has been successfully applied over rather small fully connected RNNs, on three benchmark problems of the area of control theory: pole balancing, ball throwing and octopus arm~\cite{Koutnik2010}. The authors have also applied this procedure on computer vision problems and in reinforcement learning~\cite{Koutnik2013}.
In~\cite{Butcher2013}, the authors studied the trade-off between non-linear mappings and memory capacity of ESNs. The authors applied FFTs to estimate the deviation  from linearity in the reservoir activations.
Furthermore, the state transition structure of the reservoir was analyzed in the frequency domain in~\cite{Aceituno2020}.
%

%
In our case, we also exploit the well-known benefits of Fourier transform for compressing data information.
This efficient indirect encoding allows us to represent large reservoir matrices in a reduced space using only the low frequency coefficients of the transform.
As a consequence, instead of having to tune hundreds of thousands or millions of reservoir weights, we will look for hundreds or thousands of Fourier coefficients, which implies a huge gain in complexity.
%
%
%

\medskip
\noindent\textbf{Discrete Cosine Transforms (DCTs).}
%
Let $s = (s_0, s_1, \ldots, s_{J-1})$ be a sequence of real numbers having size~$J$.
In the sequel, when we say DCT we mean the normalized DCT transform of type-II. 
The DCT $S$ of $s$ (we write $S = \text{DCT}(s)$) is also a sequence of~$J$ reals $S = (S_0, S_1, \ldots, S_{J-1})$, defined as follows:
    $$S_0 = \frac{1}{\sqrt{J}} \sum_{j=0}^{J-1} s_j,
    $$
and for $\ell = 1, 2, \ldots, J-1$,
    $$S_{\ell} = \sqrt{\frac{2}{J}}
        \sum_{j=0}^{J-1} s_j \cos\bigg(\frac{\pi \ell (2j + 1)}{2J}\bigg). 
    $$
The inverse transform, going from $S$ to $s$, denoted IDCT, is given by the following expression: for $\ell = 0, 1, \ldots, J-1$,
    $$s_{\ell} = \frac{S_0}{\sqrt{J}} 
        + \sqrt{\frac{2}{J}} \sum_{j=1}^{J-1} S_j \cos\bigg(\frac{\pi j (2\ell + 1)}{2J}\bigg). 
    $$
This IDCT transform is also called normalized DCT of type-III.
So, $S = \text{DCT}(s)$ and $s = \text{IDCT}(S)$.

DCT satisfies some properties that make the method suitable for encoding large recurrent structures and performing GAs. 
Observe that both mappings are linear, so they are continuous. Also, observe that the fact that one is inverse of the other means that they define one-to-one mappings from the space of vectors having size~$J$ to itself.

The DCT transform has a nice ``energy compaction'' property, informally meaning that if we keep in $S = \text{DCT}(s)$ just a first few terms (called ``low frequency'' terms) and replace the rest by 0s, obtaining a new vector~$S'$, then, reconstructing the initial vector from~$S'$ using the IDCT has a small error. That is, $\text{IDCT}(S') \approx s$. This characteristic of the DCT is strongly used in signal compression, and is a key element of our approach in this paper.

\subsection{Evolving reservoirs in the Fourier space}
\label{EvoESN}
%
Recall that in the canonical ESN, we select a subset of weights in the reservoir to be randomly initialized and that the rest is set to~0.
We denoted by~$N$ the number of neurons in the reservoir; let~$M$ be the number of weights randomly chosen, $M < N^2$ (for instance, $M \approx 0.2 N^2$~\cite{Jaeger09,MantasPracticalGuide12}).

%
%
%
%
Let us denote by~$\bm{p}$ the positions of the sampled weights in the reservoir matrix $\wre$, a vector having size~$M$. In our approach, we will keep fixed the vector~$\bm{p}$ but we will modify the corresponding weights during the procedure. For that reason, we call these weights the ``unfrozen'' ones, and we call the remaining $N^2 - M$ weights the frozen-to-zero ones.
%
%
After choosing an arbitrary order on those unfrozen weights, the $k$th coordinate in $\bm{p}$ corresponds to some element $(i,j)$ of $\wre$, that is, $p_k = (i,j)$. 
%
Define now vector~$\bm{\nu}$ having dimension~$M$, containing the value of those weights, that is, $\nu_k=\wre_{p_k}=\wre_{i,j}$.
%
%
%
The proposed method moves the vector $\bm{\nu}$ to the frequency domain using the DCT, and there, a GA will produce new vectors, always in the frequency space.
Call $\cal P$ this population of candidates. Each vector in $\cal P$ (denoted by $\bm{\alpha}$) will be transported to the network weights using the IDCT, occupying the fixed positions given in~$\bm{p}$. 
The ESN training procedure (e.g. ridge regression over the readout structure) will then be run, and the resulting network evaluated using a validation data set. 
The training can be done using the techniques already mentioned in Section~\ref{RC}.
This will allow us to select the best individuals of $\cal P$. 
If its quality is not good enough, the best elements in $\cal P$ are used to start a new generation of candidates through GAs.
The selection procedure used to find the best element in $\cal P$ must move each $C$-dimensional vector $\bm{\alpha}$ to the weights space, in order to evaluate the corresponding ESN. For this purpose, we first extend $\bm{\alpha}$ with zeros, to obtain a vector with size~$M$.
This procedure is called \textit{padding}, which is necessary if $C<M$ because the input and output of an inverse DCT have equal dimensions.
The dimension $C$ of the frequency space is a control parameter. The “energy compaction” property of DCTs makes that using some~$C$ value not too small, those coefficients will have enough information about the corresponding weights vectors, in spite of having much less components.

\begin{figure}[h]
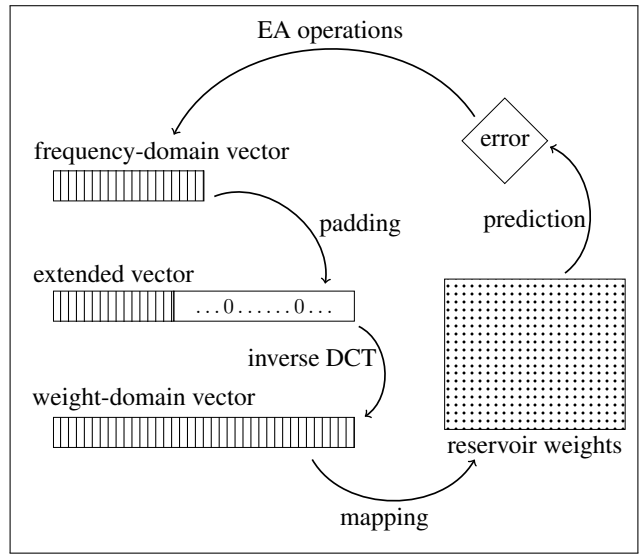

\begin{center}
\fbox{
\scalebox{0.8}{\tikz{
\tikzstyle{decision} = [diamond, draw, fill=blue!10,  font=\small, text width=4.5em, text badly centered, inner sep=0pt]
\tikzstyle{diam} = [diamond, draw, font=\small, text width=3.5em, text badly centered, inner sep=0pt]
\draw[pattern=vertical lines] (0,1.5) rectangle (2.5,1);
\node (p0) at (1.8,1.8) {\large{frequency-domain vector}};
\draw[pattern=vertical lines] (0,-0.5) rectangle (2,-1);
\node (p1) at (3.5,-0.75) {$\ldots 0 \ldots \ldots 0 \ldots$};
\draw (2,-0.5) rectangle (5,-1);
\node (p2) at (1,-0.2) {\large{extended vector}};
\node (p3) at (1.5,-2.3) {\large{weight-domain vector}};
\draw[pattern=vertical lines,] (0,-2.6) rectangle (5,-3.1);
\node (p2) at (8,-3.1) {\large{reservoir weights}};
\draw[pattern=dots] (6.5,-2.8) rectangle (9.5,-0.3);
\node (Weight1) at (5.5,-2.4) {$\;$};
\node (Res) at (6.5,-4.6) {$\;$};
\node (op1) at (5.1,0.6) {\large{padding}};
\node (op2) at (4.3,-1.6) {\large{inverse DCT}};
\node (op3) at (5.5,-4.3) {\large{mapping}};
\node (op3) at (4.6,3.8) {\large{EA operations}};
\node (r1) at (8,0.66) {\large{prediction}};
\node [diam] at  (7.5,2) {\large{error}};
\node (chromo) at (2.5,1) {$\;$};
\node (Echromo0) at (4.5,-0.5) {$\;$};
\draw (chromo) edge [->,thick, bend left=60] (Echromo0);
\node (Echromo1) at (4.9,-1) {$\;$};
\node (Weight0) at (5,-2.7) {$\;$};
\draw (Echromo1) edge [->,thick,bend left=60] (Weight0);
\draw (8.5,-0.2) edge [->,thick,bend right=60] (8.25,1.9); 
\draw (7,2.4) edge [->,thick,bend right=60] (2,2.1); 
\draw (4.3,-3.3) edge [->,thick,bend right=60] (7,-3.3); 
}}}
\end{center}
\caption{High-level visualization of the EvoESN model.}
\label{Chromo}
\end{figure}

\indent Figure~\ref{Chromo} visualizes the global EvoESN procedure. Summarizing, the padding operation is performed if necessary (if $C<M$), then sent to the weight space as a new feasible solution through the inverse DCT.
The ESN is evaluated, after training, and its error is used in the evolutionary operations.
Aiming to increase the readability, we describe the proposed algorithm into two pseudo-codes.
The evolutionary search is presented in pseudo-code~\ref{Algo1}, and the computation of the fitness value of each feasible solution by the evolutionary algorithm is given in pseudo-code~\ref{Algo2}.
Notice that in~\ref{Algo2}, the chromosome information is first transformed into the original space of weights (through $\bm{\nu} = \idct(\bm{\coeff})$), and then, before training, it is possible to re-scale the reservoir weights (in order to control the ESP). The operation of re-scaling may be optional; some authors make the spectral radius of $\wre$ larger than 1, even if this doesn't guarantee the ESP~\cite{Verstraeten07,Ferreira2013,Schrauwen07,Butcher2013}.
%
%
%
%

\section{Experiments}\label{ExperimentalResults}
To measure the computational power of the presented approach, we have
selected three well-known benchmark problems: Mackey-Glass system, Lorenz system and monthly sunspot index data.
Due to their interesting properties both chaotic systems have been largely studied in the RC community~\cite{JaegerScience04,Schmidhuber2007,JaegerConceptors,Rodan11,JaegerConceptors2018,Gallicchio2011,Butcher2013}.
In addition, monthly Sunspot data has information of the magnetic field in the Sun's outer regions.
It also has been analyzed using RC models in at least~\cite{Rodan11,Qianli2020,Sun2021}. 
An open source monthly sunspot series is available at~\cite{sidc}.
%
%
%
\subsection{Benchmark problems}
%
\noindent\textbf{The Mackey-Glass system (MGS)}.
As far as we know, ESN is still the reference RNN for modeling the MGS chaotic time series~\cite{JaegerScience04,Schmidhuber2007}, defined by the equation:
$$
\ds{\frac{{\partial y}}{\partial t}}=\ds{\frac{\alpha y(t-\tau)}{1+y(t-\tau)^{\beta}}-\gamma y(t)},
$$
where the parameters are usually $\alpha=0.2$, $\beta=10$, $\gamma=0.1$.
For these parameters' values, the system has a chaotic attractor if the delay $\tau$ satisfies~$\tau > 16.8$~\cite{JaegerScience04}. 
We study the system with a delay~$\tau=17$, the most commonly used value. 
For this problem we replicate the experimental setting used in~\cite{JaegerScience04}.
The first $1000$ time steps were discarded to washout the initial transients. 
The training length was~$3000$ time steps. The remaining~$2084$ time steps were used for testing. From those, the first 2000 steps were used in a teacher-forced mode, then the network was in exploitation mode.

\medskip
\noindent\textbf{Lorenz attractor}. The sequence data set is based on the following equations:
$$\quad
    \frac{\partial x}{\partial t} = \sigma(y-x), \quad
    \frac{\partial y}{\partial t} = rx - y - xz, \quad
    \frac{\partial z}{\partial t} = xy - bz.
$$
The parameters used are $r=28$, $b=8/3$,~$\sigma=10$, and the step size is~$0.01$. 
%
%
We replicated the pre-processing procedure for sequence $x$ and the experimental setting described in~\cite{JaegerScience04}.
The sequence was rescaled by a factor of~$0.01$, and the results are presented with data scaled back. 
There was a washout of 1000 time steps, and the training data size had~$6000$ time steps. For testing, we used~$1000$ points in a teacher-forced mode and~$600$ time steps in exploitation mode. 

\medskip
\noindent\textbf{Monthly sunspot series}. 
We studied the sunspot index from Jan. 1749 to Dec. 2021, a total of 3276 data points, available at~\cite{sidc}.
In this problem, we replicated the experimental setting investigated  in~\cite{Rodan11}. 
With a minor difference that our sequence is slightly larger than the series studied in~\cite{Rodan11}.
Initial washout of 100 samples, 1600 points for training, 500 points of validation and rest of the sequence for testing.

%
%
%


\newcommand{\vnu}{\bm{\nu}}
\newcommand{\vcoeff}{\bm{\coeff}}
%
\begin{algorithm}[ht]
\SetAlgoLined
 \tcp{Initialization of $I$ individuals in GA population}
 Random initialization of~${\cal P} = \{ \vcoeff^{(1)},\ldots,\vcoeff^{(I)} \}$\;
 Compute fitness for each $\vcoeff^{(i)}$ using Algorithm~\ref{Algo2}\;
 \tcp{Apply evolutionary search}
 \While{convergence criterion is not satisfied}{
    Selection of best individuals\;
    Mutation operation\;
    Crossover operation\;
    Compute fitness for each $\vcoeff^{(i)}$ using Algorithm~\ref{Algo2}\;
  }
 \caption{Evolutionary search for reservoir weights in the Fourier space.}
\label{Algo1}
\end{algorithm}
\begin{algorithm}[ht]
\SetAlgoLined
\If{C < M}{
    Padding operation for~$\vcoeff$\;
}
\tcp{Mapping from frequency domain to weight domain}
$\vnu = \text{IDCT}(\vcoeff)$\;
\tcp{Evaluate each individual}
Assign $\vnu$ values to  according to positions in $\bm{p}$\;
Optional scaling of $\wre$\;  
Train readout weights\;
Return ESN evaluation error of the given individual\;
\caption{Evaluation of the C-vector $\vcoeff$ (frequency-domain vector).}
\label{Algo2}
\end{algorithm}
\subsection{Hyper-parameter selection and experimental setup}
The selection of the most important hyper-parameters of the proposed EvoESN model was made following a grid search strategy.
In the case of the ESNs we also used suggestions from the literature that are described below according to each specific benchmark problem.
%
For all the benchmark problems we applied a grid search in the interval $[0.1,1.4]$ for finding a good spectral radius value.
Note that we consider spectral radius also larger than 1. As a consequence, the ESP is not guaranteed and the reservoir can exhibit unstable dynamics. This strategy was also applied in~\cite{Ferreira2013,Schrauwen07,Butcher2013}.
In the MGS and Lorenz problems some parameters, such as injected noise, input bias, activation functions, and scaling feedback connections, were directly taken from the seminal article~\cite{JaegerScience04}.

\begin{figure}[ht]
\centering
\includegraphics[scale=0.45]{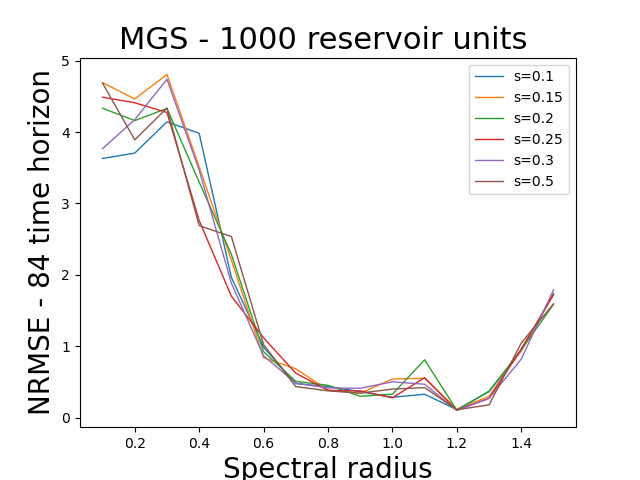}
\includegraphics[scale=0.45]{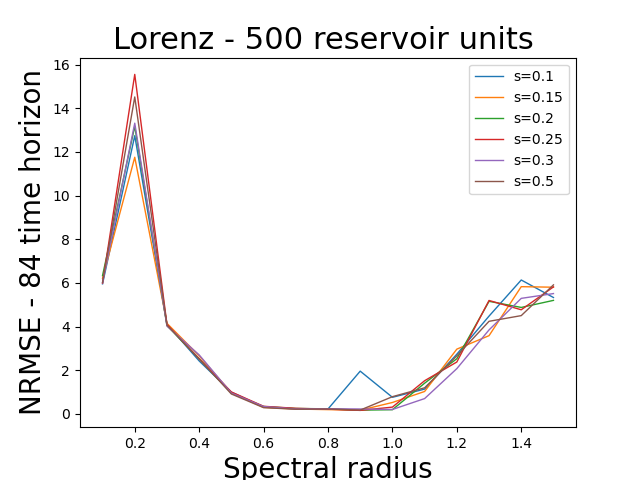}
\caption{Impact of spectral radius versus reservoir sparsity.}
\label{MGSLorenzHyperparameters}
\end{figure}

\medskip
\noindent\textbf{MGS problem.}
%
%
We adopted the network configuration employed  in~\cite{JaegerScience04}.
We designed a reservoir with~$1000$ units and the hyperbolic tangent activation function. 
%
%
The three weight matrices $\wi$, $\wre$ and $\wfb$ were sampled from the uniform distribution on~$(-1,1)$.
According to our results, a good candidate for the spectral radius of the reservoir matrix is~$0.8$, a result coherent with~\cite{JaegerScience04}. 
Left figure in Fig.~\ref{MGSLorenzHyperparameters} shows the relationship between spectral radius and reservoir sparsity. 
The curves were produced with the average of testing errors computed over 30 independent simulations.
Last, the following parameters were taken from~\cite{JaegerScience04}: injected uniform noise was scaled by $10^{-10}$, and input bias had value~$0.2$.

\medskip
\noindent\textbf{Lorenz problem.} 
For this task we also followed the hyper-parameters selection studied in~\cite{JaegerScience04}.
Input scaling and data normalization was done following~\cite{JaegerScience04}.
We used a reservoir with~$600$ units, with the hyperbolic tangent activation function. 
Right figure in Fig.\ref{MGSLorenzHyperparameters} shows the testing error for different values of reservoir spectral radius and reservoir sparsity.
Curves were created averaging the results of 30 independent simulations.
We observed that the best spectral radius candidates were close to~$1$. 
The reservoir matrix had~$20\%$ of non-zero weights. 
Then, we set up a spectral radius to~$0.97$ (the same value was selected in~\cite{JaegerScience04}).
The input and reservoir matrices were initialized with the uniform distribution over~$(-1,1)$. Uniform injected noise had values~$\pm 10^{-7}$~\cite{JaegerScience04}.
The feedback connections were sampled following an uniform distribution in~$(-4,4)$ (this particular range was suggested in~\cite{JaegerScience04}).

\medskip
\noindent\textbf{Sunspot series problem.}
We adopted the setting used in~\cite{Rodan11}, and we also analyzed some parameters using grid search. 
An important difference in this problem with the previous task, is that the goal is to predict only one single time-step ahead (as it was studied in~\cite{Rodan11}). 
In previous tasks we applied free run prediction over large testing time windows.
For this task we studied only networks without feedback connections. Left figure in Fig.~\ref{MG-C} shows the relationship between spectral radius and sparsity of the reservoir.
Besides, for this problem we also analyzed leaky units in the range $[0.01:0.05:0.5]$. For the regularization parameter we evaluated the values $10^{-i}$ with $i\in\{3,4,5,6,7,8\}$, and selected $10^{-5}$. The selected input scaling was $0.1$ after evaluation over the grid  $\{0.01,0.02,0.03,0.05,0.07,0.09,0.1,0.15,0.2,0.3,0.5\}$.

\medskip
\noindent\textbf{Computational aspects.}
The IDCT was chosen with the orthogonal norm using the \texttt{fftpack Python} package.
We conducted GA as EA, and it was implemented using the \texttt{DEAP} package~\cite{DEAPJMLR2012}. 
We analyzed the impact of $C$ (number of coefficients in the frequency-domain vector) in a size range between~$25$ and~$750$.
Besides, different numbers of generations were evaluated:~$70$,~$150$,~$300$,~$600$. 
After carrying out multiple runs of the GA, we selected the hyper-parameters in order to have relatively fast convergence and to diminish the probability of being trapped in a local minima.
%
Here we present the results for the following GA parameters. We used a Gaussian mutation with parameter~mutpb=0.15, two-point crossover, probability of mating two individuals of $0.5$, number of individuals participating in each tournament~$= 3$, and population size~$= 20$. 
%
%
%
The readout was computed using a ridge regression with a regularization factor of~$10^{-9}$ (MGS task) and~$10^{-6}$ (Lorenz task)~\cite{JaegerScience04}, and $10^{-5}$ (Sunspot task).
%

%
%

%
\subsection{Results analysis}
%
%
%
\noindent\textbf{Accuracy evaluation.} It is common to see the NRMSE metric on experiments conducted with MGS and Lorenz data~\cite{JaegerScience04,Schmidhuber2007,Butcher2013}. Therefore, for comparison purposes we also used NRMSE for these two tasks, and for the sunspot series problem we use NMSE as it was applied in~\cite{Rodan11}.
%
%
%
We denote by NMRSE($H$) the error computed in the testing phase using the residuals over \emph{the whole horizon}~$H$ (by residual we mean the difference between target and predicted values).
We also present another metric denoted by~NRMSE$_{H}$ following the notation in~\cite{JaegerScience04,Schmidhuber2007}.
NRMSE$_{H}$ is computed in the testing phase after~$H$ autonomous steps. The error is computed only using the residual at the~$H$ time step.
NRMSE$_{84}$ has become the standard metric for evaluating the MGS prediction~\cite{Schmidhuber2007}.

\medskip
\noindent\textbf{Accuracy comparison.}
Table~\ref{Literature} presents a non-exhaustive overview of the reported state-of-the-art results. Previous studies have analyzed the problems using different error metrics.
Then, the last column in the table shows the information about the error measure.
Furthermore, we also implemented the canonical ESN, the ESN with leaky integrator units~\cite{Jaeger07}, and one variation of ESN with feedback connections introduced in~\cite{JaegerScience04}. 
In~\cite{JaegerScience04}, the authors describe a learning approach for correcting the bias introduced by the teacher-forcing schema in the case of feedback connections.
Jaeger \textit{et al.} obtained the best performance for MGS prediction using this particular learning schema introduced as \textit{refined learning method}. 
Let \textit{ESN with FB (I)} denote the ESN with feedback connections using the teacher-forcing schema, and \textit{ESN with FB (II)} denote the ESN with feedback connections using the refined learning method.
%
%
%
%

\begin{figure}[ht]
\centering
\includegraphics[scale=0.45]{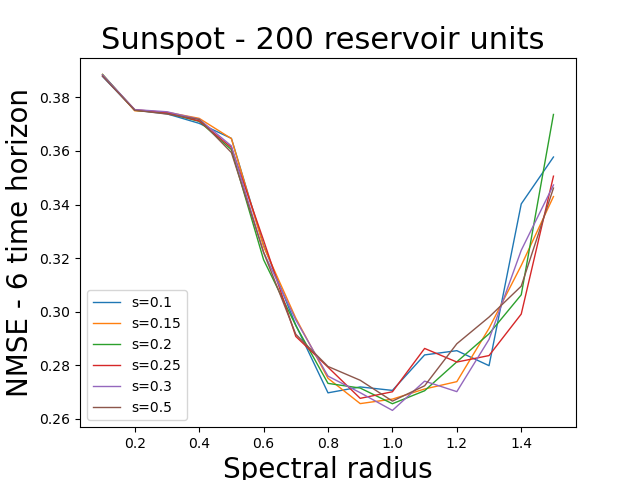}
\includegraphics[scale=0.45]{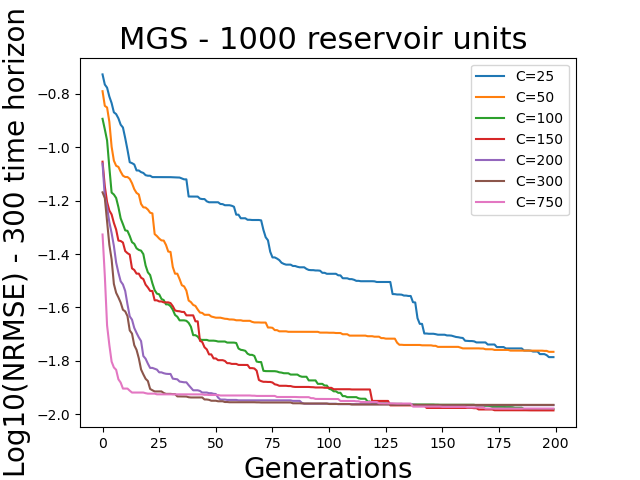}
\caption{
Impact of hyper-parameters. Left figure shows the relationship between spectral radius and sparsity in the Sunspot series problem. Right figure shows the impact of the number of coefficients $C$ in the MGS problem.}
\label{MG-C}
\end{figure}

\begin{figure}[ht]
\centering
\includegraphics[scale=0.6]{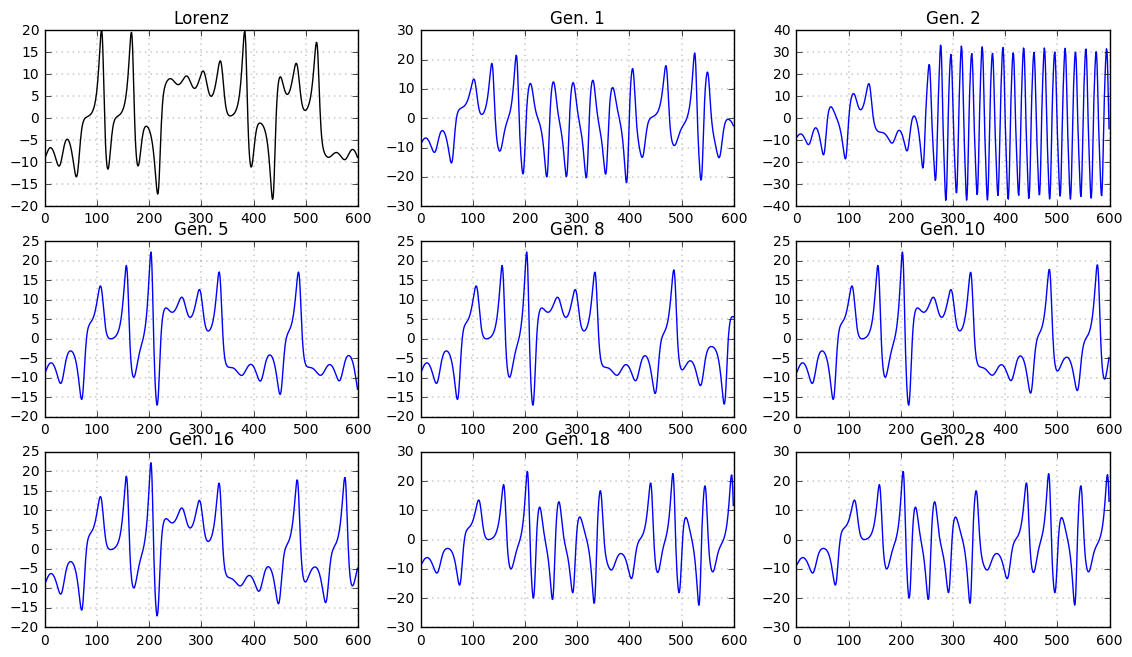}
\caption{Prediction of the Lorenz~attractor: predictions of randomly selected individual according to the GA generations.}
\label{LorenzEvoESNCoeff150}
\end{figure}

\begin{figure}[ht]
\centering
\includegraphics[scale=0.6]{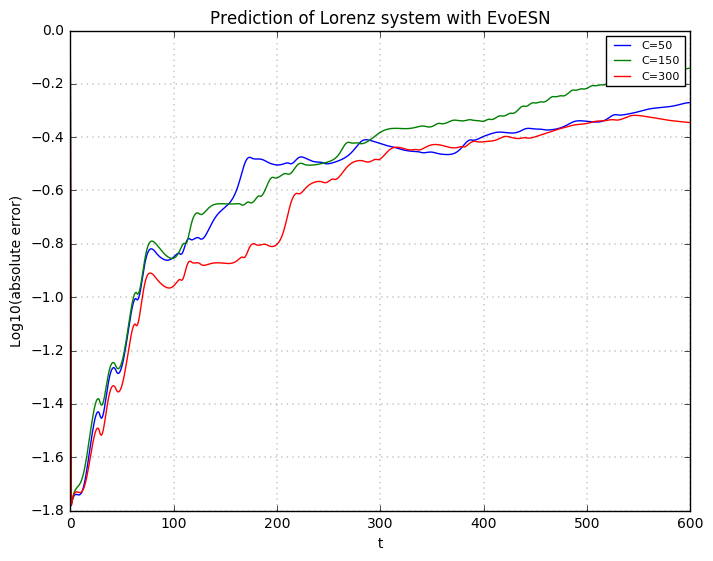}
\caption{Prediction of the Lorenz attractor for different values of $C$. Type of visualization taken from~\cite{JaegerScience04}}
\label{LorenzAbsoluteError}
\end{figure}

\medskip
\noindent\textbf{Results.}
%
We partially investigated the impact of the parameter $C$. 
%
Some results are shown in the right figure in Fig.~\ref{MG-C} where the prediction error, in a logarithmic scale, is plotted against time (generation number), for different values of~$C$. 
The fitness function was computed using 300 points of generalization on~$12$ independent prediction tasks.
The curves show the evolution of the average of NRMSE across the~$12$ independent experiments.
They also show that the accuracy is increasing with $C$ and with time.
See that if $C \geq 200$, for instance, then after generation~$50$, the performance of the method becomes roughly independent of~$C$. The results are consistent, the curves behave as expected, and globally, the procedure shows its efficiency once the frequency-domain vectors are large enough.
Figure~\ref{LorenzEvoESNCoeff150} shows results over Lorenz data with~$C = 150$.
There is a group of~$6$ figures. The top-left figure has an example of Lorenz testing data. Next graphics show EvoESN predictions (of a randomly selected individual) for different generations of the GA.
In a few iterations, the model is able to reproduce with significant accuracy a large segment of the Lorenz attractor.
Figure~\ref{LorenzAbsoluteError} shows the absolute error for different~$C$ values on~600 time units in the Lorenz system task. For the three values of $C$ the results outperform the ones reported in~\cite{JaegerScience04} using different variations of ESNs.
Note that the red curve ($C=300$) has a better accuracy than the other two curves ($C=50$ and $C=150$) until the time horizon $t \approx 300$. 
For longer time horizons than~$t=300$ both the red and blue curves seem to have similar behavior.
The explanation probably lies in that the fitness function used in the GA algorithm was computed using 200 points of generalization, i.e. the reservoir weights are optimized for predicting the next 200 points. After that threshold limit the models behaves similarly (independently of $C$).
Further possible improvements will be looked for by studying different fitness functions in the evolutionary search.

Table~\ref{ResultsMG} summarizes the results of experiments over MGS.
In the case of EvoESN, we present the results for $500$ coefficients, and~$70$ and $300$~generations in the GA algorithm. 
A total of~50 randomly created ESNs were evaluated. We show the average of the error produced by independent runs.
EvoESN reaches much lower accuracy than the obtained with ESNs using our implementation.
EvoESN also reaches the results reported in~\cite{JaegerScience04}. 
However, even though the variance is very low, we cannot say that the difference is significant with the error reported in~\cite{JaegerScience04}.
We expect to do further studies using statistical analysis and a full evaluation of the GA parameters, in order to check if the difference is statistically significant.

Table~\ref{Lorenz} shows the results for the Lorenz problem.  
The NRMSE$_{84}$ error has the same order of magnitude as obtained by Jaeger \textit{et al.} using a feedback connection and a refined learning method~\cite{JaegerScience04}.
It also shows the absolute error over a~600 time horizon, that was studied in~\cite{JaegerScience04}.
Table~\ref{TableSunspot} presents the test errors for the Sunspot series problem. EvoESN outperforms the results reported in~\cite{Rodan11}, and also the results obtained by our implementations of ESN with leaky-integrators. We ran 50 independent simulations with the same ESN architecture, and the standard deviation of error across simulation has been really low ($\approx1\times 10^{-6}$ independently of the leaky rate. Standard deviation values for EvoESN experiments are shown in the table.
%
%
Regarding complexity, EvoESN has a higher cost than standard ESNs. Note that the GA performs, at each generation, training of the readouts using linear regression, computation of reservoir spectral radius and of inverse DCT. Other GA operations such as crossover, selection and mutation are less costly than the mentioned above.
However, it seems that a few iterations are enough for achieving much lower errors with respect to ESNs.
%
%
%
%
%
\newpage
\section{Conclusions}\label{Conclusions}
%
%
In this article, we have presented the EvoESN, an RC-based model where the reservoir weights are not set randomly, but are obtained by applying GA, integrated with the classical training of the ESN readouts.
The proposed algorithm projects the reservoir weights into the frequency domain using a DCT.
The optimization is performed in a frequency space whose dimension is much smaller than that of the non-null reservoir weights.
This leads to a significant dimensionality reduction, in turn producing a significant computer time reduction, while keeping the quality of the optimization procedure. These are consequences of the properties of Fourier-like transforms.
Since we use an indirect encoding the dimension of the searching space is basically independent of the reservoir size, an additional advantage of the proposed technique. Another advantage is that the fine-tuning of the reservoir weights is made using the fitness function of the evolutionary search, which can be different than the one used for training the readouts.
In addition, by applying GA we avoid the difficulties associated with the training of RNNs. 
The quality of these first experiments clearly show the potential of the approach, and the first work to be done in the close future is to push the procedure towards better performances by a careful analysis of both weight space encoding and evolutionary search procedures. 

\subsection*{Acknowledgements}
This work was supported by the GACR-Czech Science Foundation project no. 21-33574K ``Lifelong Machine Learning on Data Streams''.

\begin{table}[ht]
\caption{\label{Literature}Accuracy comparison with other RC methods.}
\centering
\scalebox{0.95}{
 \begin{tabular}{llccc}
 \toprule
 \toprule
Problem & Reference &  Model & Error & Metric\\
    \toprule
\multirow{5}{*}{MGS} &  H. Jaeger~\textit{et al.}~\cite{JaegerScience04} & ESN-FB & $\approx 10^{-4.2}$  & NRMSE$_{84}$\\ 
  ~ & Schmidh\"{u}ber~\textit{et al.}~\cite{Schmidhuber2007}   & Evolino & $\approx 1.9\times10^{-3}$ &  NRMSE$_{84}$\\   
~ & J. Steil~\cite{Steil04} & BPDC & 0.318 & NMSE\\
~& A. Ferreira \textit{et al.}~\cite{Ferreira2013} & RCDESIGN$\:$ & $0.00051$ & NRMSE\\
~& C. Gallicchio \textit{et al.}~\cite{Gallicchio2011} & RDESN & $1.112\times 10^{-8}$ & MSE\\
\midrule
   \multirow{1}{*}{Lorenz} & Jaeger~\textit{et al.}~\cite{JaegerScience04} & ESN-FB & $\approx -4.27$  & $log_{10}($NRMSE$_{84}$)\\ 
\midrule 
\multirow{2}{*}{Sunspot} & Rodan~\textit{et al.}~\cite{Rodan11} & ESN & $\approx 0.1042$  & NMSE\\
~ & Rodan~\textit{et al.}~\cite{Rodan11} & \mbox{DLR, DLRB, SCR} & $\approx 0.1039$  & NMSE\\
\bottomrule
\bottomrule
\end{tabular}
}
\vspace{0.5in}
%
 \caption{ \label{ResultsMG}Comparison between different types of ESNs and EvoESN (with $C=500$) for the MSG prediction task. 
 }
\centering
\scalebox{1}{
 \begin{tabular}{p{1.5cm}p{4.5cm}p{2.9cm}p{2.5cm}}
\toprule
\toprule
Model & Characteristics & Error & Metric \\
    \toprule
%
ESN & without FB, non-linear readout &  $0.1770$ & NRMSE ($84H$)\\ 
ESN & without FB, linear readout &  $0.29857$ & NRMSE ($84H$)\\
ESN & with FB (I), linear readout & $0.19620$ & NRMSE ($84H$)\\
ESN & with FB (I), non-linear readout & $0.19376$ & NRMSE ($84H$)\\
ESN & with FB (II), non-linear readout & $0.07684$ & NRMSE ($84H$)\\
\midrule
EvoESN & $300$ generations & $~\approx 1.075 \times 10^{-4}$ & NRMSE$_{84}$\\
EvoESN & $70$ generations & $4.485 \times10^{-4}$ & NRMSE ($84H$)\\
\bottomrule
\bottomrule
\end{tabular}
}
\end{table}
\begin{table}[ht]
\begin{center}
\caption{\label{Lorenz}Results for the Lorenz attractor prediction.}
\scalebox{1}{
 \begin{tabular}{lccc}
 \toprule
  \toprule
Model & $C$ & Error & Metric \\
    \toprule
EvoESN & 50 & -3.4666 & $log_{10}$ NRMSE$_{84}$ \\
EvoESN & 100 & -3.6014 & $log_{10}$ NRMSE$_{84}$ \\
EvoESN & 150 & -4.1379 & $log_{10}$ NRMSE$_{84}$ \\
\midrule
EvoESN & 50 & -0.2704 & $log_{10}$ absolute error (600H)\\ %
EvoESN & 150 & -0.1407 & $log_{10}$ absolute error (600H)\\ 
\bottomrule
\bottomrule
\end{tabular}
}
\end{center}
\end{table}
%
%
\begin{table}[ht]
\begin{center}
\caption{\label{TableSunspot}Results for the Sunspot prediction.}
\scalebox{1}{
 \begin{tabular}{llcc}
 \toprule
  \toprule
Model & Characteristics & Error & Metric \\
    \toprule
%
ESN & without FB  & 0.085358 & NMSE\\
LI-ESN & leaky rate 0.1 & 0.033120 & NMSE \\
LI-ESN & leaky rate 0.7 & 0.031816 & NMSE \\
\midrule
EvoESN & C 50, NGEN 150 & 0.0290556 ($8.1\times 10^{-4}$) & NMSE \\
EvoESN & C 75, NGEN 150 & 0.0291978 ($1.9\times 10^{-4}$) & NMSE \\
EvoESN & C 100, NGEN 150 & 0.0292394 ($7.4\times 10^{-5}$) & NMSE \\
EvoESN & C 200, NGEN 150 & 0.0293263 ($7.7\times 10^{-5}$) & NMSE \\
EvoESN & C 300, NGEN 150 & 0.0309722 ($4.4\times 10^{-5}$) & NMSE \\
\bottomrule
\bottomrule
\end{tabular}
}
\end{center}
\end{table}
\clearpage

\bibliographystyle{plain}
\bibliography{References}






\end{document}